%% file: egpaper_final.tex
\pdfoutput=1
\documentclass[10pt,twocolumn,letterpaper]{article}

\usepackage{cvpr}
\usepackage{times}
\usepackage{epsfig}
\usepackage{graphicx}
\usepackage{amsmath}
\usepackage{amssymb}

\usepackage{comment}

\cvprfinalcopy 

\def\cvprPaperID{1948} 

\ifcvprfinal\fi
\begin{document}

\title{SCT: Set Constrained Temporal Transformer for Set Supervised Action Segmentation}

\author{
Mohsen Fayyaz and Juergen Gall\\
University of Bonn\\
Bonn, Germany\\
{\tt\small \{fayyaz, gall\}@iai.uni-bonn.de}
}

\maketitle

\begin{abstract}
\input{Paper/0Abstract.tex}
\end{abstract}

\input{Paper/1Introduction.tex}%

\input{Paper/2RelatedWork.tex}%

\input{Paper/3ActionSet.tex}%

\input{Paper/4ProposedMethodArchitecture.tex}

\input{Paper/5ProposedMethodTraining.tex}%

\input{Paper/6Experiments.tex}%

\input{Paper/7Concolusion.tex}

\input{Paper/8Acknowledgment.tex}

{\small
\bibliographystyle{ieee_fullname}
\bibliography{references}
}

\end{document}

%% file: Paper/0Abstract.tex
Temporal action segmentation is a topic of increasing interest, however, annotating each frame in a video is cumbersome and costly. Weakly supervised approaches therefore aim at learning temporal action segmentation from videos that are only weakly labeled. In this work, we assume that for each training video only the list of actions is given that occur in the video, but not when, how often, and in which order they occur. In order to address this task, we propose an approach that can be trained end-to-end on such data. The approach divides the video into smaller temporal regions and predicts for each region the action label and its length. In addition, the network estimates the action labels for each frame. By measuring how consistent the frame-wise predictions are with respect to the temporal regions and the annotated action labels, the network learns to divide a video into class-consistent regions. We evaluate our approach on three datasets where the approach achieves state-of-the-art results.

%% file: Paper/1Introduction.tex
\section{Introduction}

For many applications large amount of video data needs to be analyzed. This includes temporal action segmentation, which requires to label each frame in a long video by an action class. In the last years, several strong models for temporal action segmentation have been proposed~\cite{kuehne2016end, lea2017temporal, MS-TCN}. These models are, however, trained in a fully supervised setting, \ie, each training video needs to be fully annotated by frame-wise labels. Since acquiring such annotations is very expensive, several works investigated methods to learn the models with less supervision. An example of weakly annotated training data are videos where only transcripts are provided~\cite{hildecviu, ectc, richard2017weakly, richard2018nnviterbi, isba, d3tw, MuCon, CDFL}. While transcripts of videos can be obtained from scripts or subtitles, they are still costly to obtain. In \cite{ActionSet} it was therefore proposed to learn temporal action segmentation only from a set of action labels that are provided for a complete video of several minutes. In this case, it is only known which actions occur, but not when, in which order, or how often. This makes the task much more challenging compared to learning from transcripts or fully supervised learning.        

In the work \cite{ActionSet}, the problem has been addressed by hypothesizing transcripts that contain each action label of a video at least once and then infer a frame-wise labeling of the video by aligning the hypothesized transcripts. While the approach showed that it is possible to learn from such weak annotation even for long videos, the approach does not solve the problem directly but converts it into a weakly supervised learning problem where multiple hypothesized transcripts per video are given. This is, however, ineffective since it is infeasible to align all transcripts that can be generated from a set of action labels and it uses the provided annotations not directly for learning.                 

In this work, we propose a method that uses the action labels that are given for each training video directly for the loss function. In this way, we can train the model in an end-to-end fashion. The main idea is to divide a video into smaller temporal regions as illustrated in Figure \ref{fig:temporalRegionsLengths}. For each region, we estimate its length and the corresponding action label. Since for each training video the set of actions is known, we can directly apply a set loss to the predicted action labels of the temporal regions, which penalizes the network if it predicts actions that are not present in the video or if it misses an action. The problem, however, is that we cannot  directly apply a loss to the prediction of the region lengths. While a regularizer for the predicted length that penalizes if the lengths of the regions get too large improves the results, it is insufficient as we show in our experiments. 
We therefore introduce a second branch to make frame-wise predictions and measure how consistent the frame-wise predictions are with respect to the temporal regions and the annotated action labels.
Using our differentiable Set Constrained Temporal Transformation (SCT), this loss affects the lengths of the regions, which substantially improves the accuracy of the model.

In our experimental evaluation on three datasets, we show that the proposed approach achieves state-of-the-art results. We furthermore thoroughly evaluate the impact of each component.

%% file: Paper/2RelatedWork.tex
\section{Related Work}

Researchers in the field of action recognition have made significant advances in recent years. Methods for action recognition on trimmed video clips have acquired prominent achievements in recent years~\cite{i3d, STC-Net, T3D, HVU, nonlocal, slowfast, Tran_2019_ICCV}. Although current methods achieve high accuracies on large datasets such as Kinetics~\cite{kinetics}, HMDB-51~\cite{hmdbdataset}, and UCF-101~\cite{ucfdataset}, in realistic problems videos are not temporally trimmed. 

Using the publicly available untrimmed video action segmentation datasets such as Breakfast~\cite{BreakFastDataset} or ActivityNet~\cite{ActivityNet}, several works address action segmentation in videos~\cite{kuehne2016end, lea2017temporal, zhao2017temporal, MS-TCN}.
Early action segmentation methods utilized Markov models on top of temporal models \cite{lea2016segmental, kuehne2016end} or sliding window processing \cite{rohrbach2012database, karaman2014fast}. \cite{Richard16} models context and length information. They show that length and context information significantly improve action segmentation. There are also other fully supervised methods that use grammars \cite{Pirsiavash2014,StochasticGrammar2014,HildeWACV2016}.
Recent methods try to capture the long range temporal dependencies using temporal convolutions with large receptive fields \cite{lea2017temporal, MS-TCN}.

The existing methods in weakly supervised action segmentation use ordered action sequences as annotation. The early works tried to get ordered sequences of actions from movie scripts \cite{Laptev2008, Duchenne2009}.
Bojanowski et~al.~\cite{hollywoodextended} introduced the Hollywood extended dataset. They also proposed a method for action alignment based on discriminative clustering.
Huang et~al.~\cite{ectc} proposed to use an extended version of the CTC loss. Kuehne et~al.~\cite{hildecviu} proposed a HMM-GMM based system that iteratively generates pseudo ground truth for videos during training.
Richard et~al.~\cite{richard2017weakly} use an RNN for short range temporal modeling. 
Most of these methods rely on iterative pseudo ground-truth generation approaches which does not allow for end-to-end training.
Richard et~al.~\cite{richard2018nnviterbi} introduced the Neural Network Viterbi (NNV) method. They use a global length model for actions, which is updated during training. Souri et~al.~\cite{MuCon} introduce an end-to-end method which does not use any decoding during training. They use a combination of a sequence-to-sequence model on top of a temporal convolutional network to learn the given transcript of actions while learning to temporally segment the video.
Li et~al.~\cite{CDFL} build upon NNV  which achieves state-of-the-art results in weakly supervised action segmentation with ordering constraints.

When working with weak supervision without ordering constraints, only the set of actions is given during training. Richard et~al.~\cite{ActionSet} address the problem by hypothesizing transcripts that contain each action label of a video at least once and then infer a frame-wise labeling of the video by aligning the hypothesized transcripts. They showed that it is possible to learn from such weak annotation even for long videos, but they do not solve the problem directly. They convert the problem into a weakly supervised learning problem where multiple hypothesized transcripts per video are given. This is, however, ineffective since it is infeasible to align all transcripts that can be generated from a set of action labels and it uses the provided annotations not directly for learning. 

%% file: Paper/3ActionSet.tex
\section{Weakly Supervised Action Segmentation}

Action segmentation requires to temporally segment all frames of a given video, \ie, predicting the action in each frame of a video. The task can be formulated as follows. Given an input sequence of $D$-dimensional features $X_{1:T} = (x_1, \dots, x_T)$, $x_t \in \mathbb{R}^D$, the task is to infer the sequence of framewise action labels $\hat{Y}_{1:T}=(\hat{y}_1,\dots,\hat{y}_T)$ where there are $C$ classes $\mathcal{C}=\{1,\dots,C\}$ and $\hat{y}_{t} \in \mathcal{C}$.

In the case of fully supervised learning, the labels $\hat{Y}_{1:T}$ are provided for each training sequence. In this work, we investigate a weakly supervised setting as in \cite{ActionSet}. In this setting, only the actions $\hat{A}=\{\hat{a}_1,\ldots,\hat{a}_M\}$ that occur in a long video are given where $\hat{a}_m \in \mathcal{C}$ and $M \leq C$. In contrast to other weakly supervised settings where transcripts are given, this is a much more difficult task since not only the lengths of the actions are unknown for the training sequences, but also the order of the actions and the number of the occurrences of each action.   

%% file: Paper/4ProposedMethodArchitecture.tex
\section{Proposed Method}\label{sec:propsedMethod}

\begin{figure}[t]
    \centering
    \includegraphics[scale=0.09,bb=0 0 2667 1100]{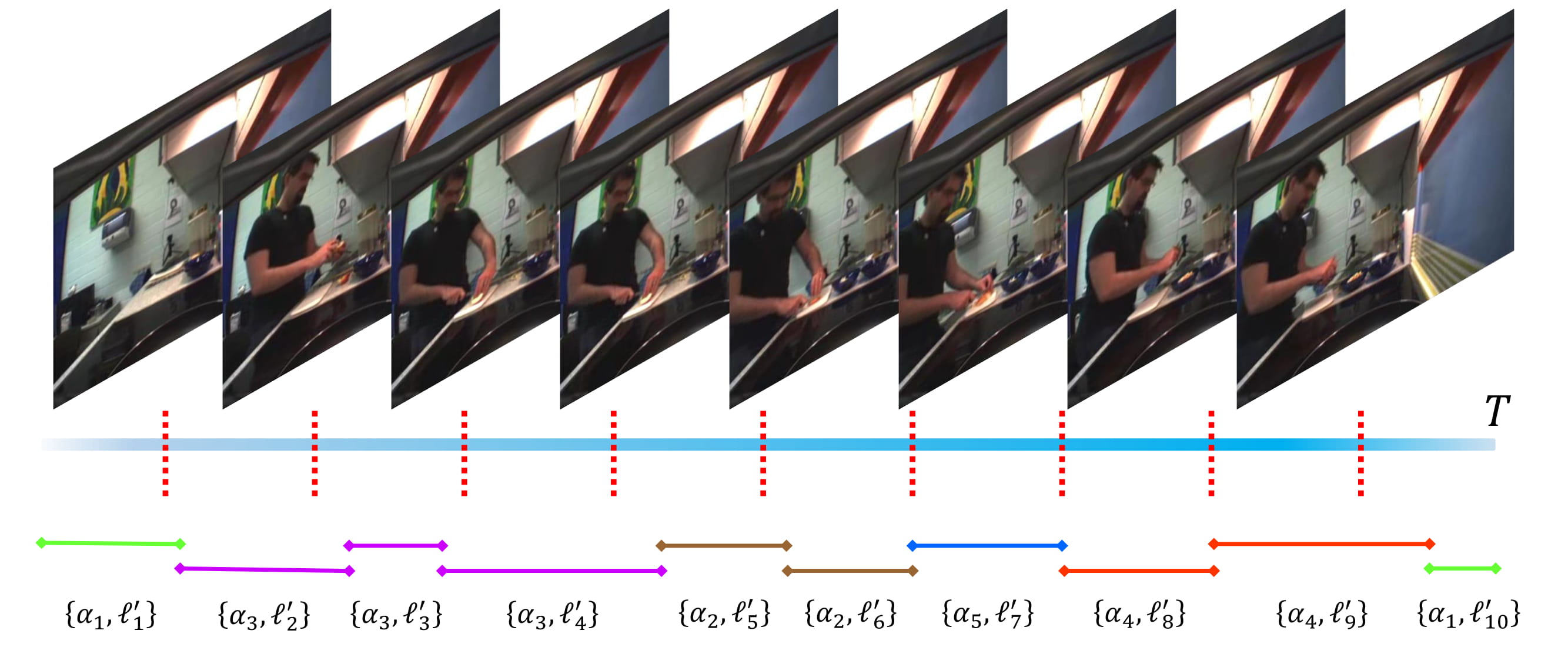}
    \caption{Our model estimates for $K$ temporal regions the actions probabilities $A_{1:K}=(a_1,\dots,a_{K}), a_k \in \mathbb{R}^{C},$ and the temporal lengths of the regions $L_{1:K}=(\ell_1,\dots,\ell_{K}), \ell_k \in \mathbb{R}$. In this example, $K=10$. Since temporal regions are not aligned with the action segments, the model estimates the temporal lengths to refine the corresponding temporal region of the predicted action. }
    \label{fig:temporalRegionsLengths}
\end{figure}

\begin{figure*}[t]
    \centering
    \includegraphics[scale=0.73,bb=0 0 650 350]{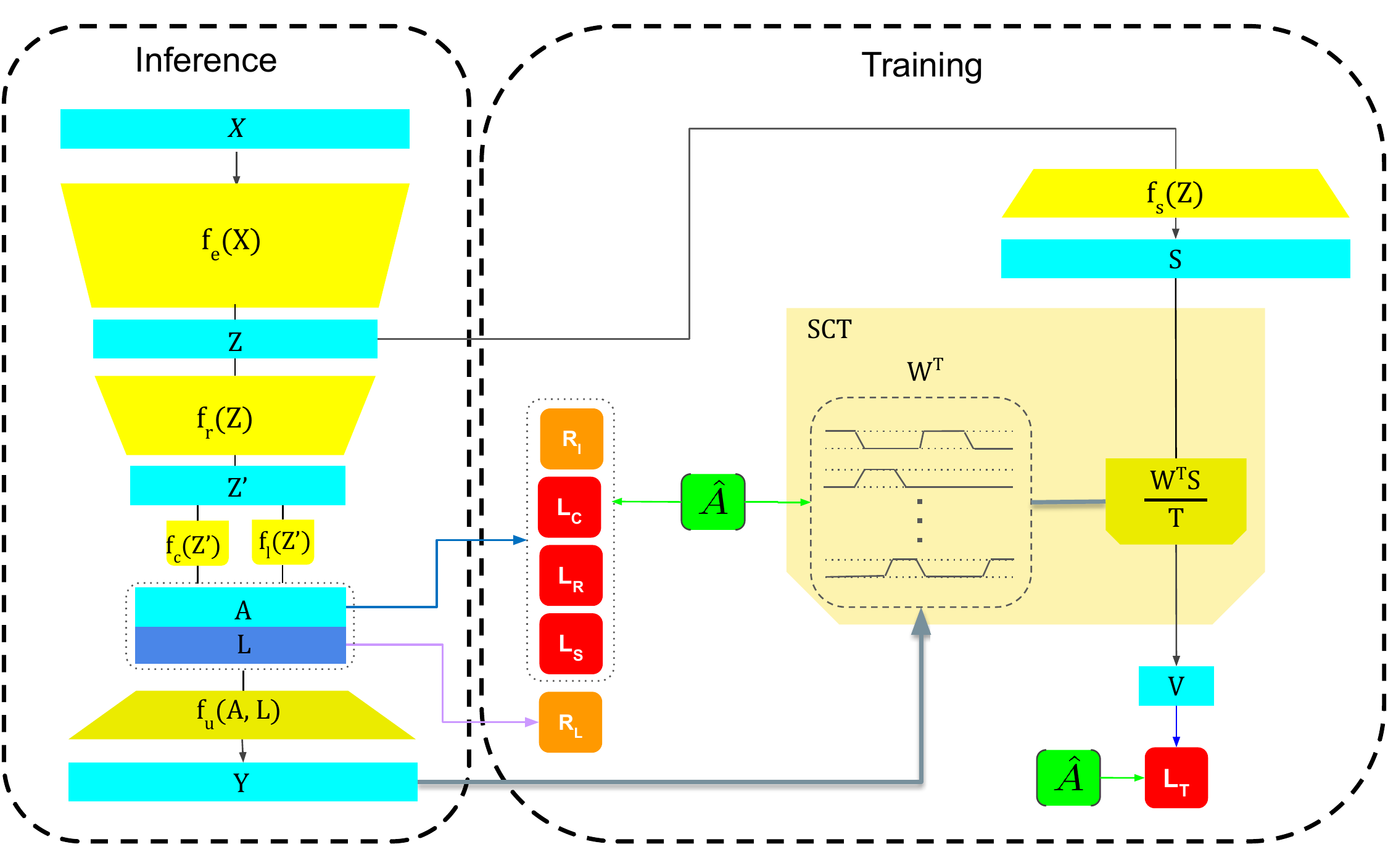}
    \caption{Overview of the proposed network with loss functions. The network gets a sequence of features $X_{1:T}$ as input. A temporal model $f_e(X)$ maps these features to a latent space $Z$ with lower temporal resolution. The lower branch $f_r(Z)$ divides the temporal sequence into temporal regions $Z'_{1:K}$ and estimates for each region the action probabilities $a_{k}$ and the length $l_{k}$. Since the temporal resolution has been decreased, the upsampling module $f_u(A,L)$ uses the lengths $L_{1:K}$ and the action probabilities $A_{1:K}$ of all regions to obtain estimates of the framewise probabilities $Y_{1:T}$. While $L_{1:K}$ is regularized by the length regularizer $\mathcal{R_L}$, $A_{1:K}$ is trained to minimize $\mathcal{L_S}$, $\mathcal{L_R}$, $\mathcal{L_C}$, and $\mathcal{R_I}$. Since besides of the regularizer $\mathcal{R_L}$, there is no loss term that provides supervision for $L$, we use a second branch $f_s(Z)$ to provide an additional supervisory signal. Using SCT, we transform the temporal representations $Y_{1:T}$ and $S_{1:T}$ to a set representation $V_{1:M}$ for the self supervision loss $\mathcal{L_T}$. 
    }
    \label{fig:overview}
\end{figure*}

In order to address weakly supervised action segmentation, we propose a network that divides a temporal sequence into temporal regions and that estimates for each region the action and the length as illustrated in Figure~\ref{fig:temporalRegionsLengths}. This representation is between a frame-wise representation where the length of each region is just one frame and an action segment representation where a region contains all neighboring frames that have the same action label.\\                 
Figure~\ref{fig:overview} illustrates our proposed network, which consists of three components. The first component $f_e(X)$, which is described in Section \ref{sec:temporalmodule}, maps the input video features $X \in \mathbb{R}^{T \times D}$ to a temporal embedding $Z \in \mathbb{R}^{T' \times D'}$ where $T' < T$ and $D' < D$. The second component $f_r(Z)$, which is described in Section \ref{sec:tempSegment}, takes $Z$ as input and estimates for $K$ temporal regions the actions probabilities $A_{1:K}=(a_1,\dots,a_{K}), a_k \in \mathbb{R}^{C},$ and the temporal lengths of the regions $L_{1:K}=(\ell_1,\dots,\ell_{K}), \ell_k \in \mathbb{R}$. In order to obtain the frame-wise class probabilities $Y \in \mathbb{R}^{T \times C}$ from $L$ and $A$, the third component $f_u(A,L)$, which is discussed in Section \ref{sec:tempTransforming}, upsamples the estimated regions such that there are $T$ regions of length $1$.

\subsection{Temporal Embedding}\label{sec:temporalmodule}
Given the input video features $X \in \mathbb{R}^{T \times D}$ the temporal embedding component $f_e(X)$ outputs the hidden video representation $Z \in \mathbb{R}^{T' \times D'}$. Our temporal embedding is a fully convolutional network. In this network we first apply a 1-d convolution with kernel size $1$ to reduce the input feature dimension from $D$ to $D'$. On top of this layer we have used $B$ temporal convolution blocks (TCB) with $B=6$. Each TCB contains a dilated 1-d convolution layer with kernel size $3$ for conducting the temporal structure. We increase the dilation rates as $\{2^b | b \in \mathbb{Z^{+}}, 0\leq b\leq B\}$ where $b$ is the index of the TCBs. Then a ReLU activation function is applied on top of the convolutional layer. On top of this combination, a 1-d convolution layer with kernel size $1$ and a residual connection is used. Finally, a dropout with a probability of $0.25$ is applied on top. The TCB is modeled after the WaveNet architecture \cite{WaveNet}. To reduce the temporal dimension of the representation, we perform temporal max poolings with a kernel size of $2$ on top of the TCBs with $b=\{1,2,4\}$. Using the TCBs and max poolings, we get large receptive fields on the input data $X$. Having such large receptive fields provides the capability of modeling long and short range temporal relations between the input frames. 

\subsection{Temporal Regions}\label{sec:tempSegment}
On top of the temporal embedding, we use the temporal region estimator network $f_r(Z)$ to estimate the action probabilities and the temporal lengths for $K$ temporal regions. $f_r(Z)$ outputs the hidden representation $Z'_{1:K}=(z'_1,\dots,z'_K), z'_k \in \mathbb{R}^{D'}$, for the temporal regions. To have a better representation for estimating the  actions probabilities $A$ and region lengths $L$, we increase the receptive field size and decrease the temporal dimension. This network mostly follows the same architecture design as $f_e(X)$. It has $B'$ TCBs with $B'=4$. The dilation rates of the TCBs are set as $\{2^{b'} | b' \in \mathbb{Z^{+}}, B< b'\leq B+B'\}$. To reduce the temporal dimension of the representation, we perform temporal max poolings with kernel size $2$ on top of the TCBs with indices 2 and 4. On top of the final TCB, we have two different heads $f_{c}$ and $f_{l}$. $f_{c}(Z')$ predicts the class probabilities $A$. It consists of a 1-d convolution layer with a kernel size of $1$ and an output channel size of $C$. A softmax function is applied on top of the convolution layer to get the action probabilities $A$. $f_{l}(Z')$ predicts the lengths $L$ for the corresponding temporal regions. It consists of two 1-d convolution layers with kernel sizes $1$ and output channels $D'/2$ and $1$, respectively. 

\subsection{Region Upsampling}\label{sec:tempTransforming}
$f_{c}(Z')$ estimates the action probabilities $A_{1:K}$ for temporal regions. To get probabilities for temporal action segmentation, we need to upsample $A_{1:K}$ to $Y_{1:T}$. Since $f_l(Z')$ predicts the corresponding lengths $L_{1:K}$, we can use theses lengths to upsample the probabilities $A$. To do so, we first project the predicted lengths $L_{1:K}$ to absolute lengths $L'_{1:K}=(\ell'_{1},...,\ell'_{K}),\ell'_{k}\in\mathbb{Z^+}$, by:
\begin{equation}
    \label{eq:length_normalize}
    \ell'_k = T  \frac{e^{\ell_k}}{\sum_{i=1}^{K} e^{\ell_i}}.
\end{equation}
In other words, we apply the softmax function on $L$ to get the relative lengths, which sum up to $1$ and then multiply them by $T$ to get the absolute lengths. Therefore, the absolute lengths sum up to $T$, which is our desired final temporal size for $Y$. Given the absolute lengths $L'$, we upsample $A$ in a differentiable way such that $a_k \in \mathbb{R}^{C}$ becomes $a'_k \in \mathbb{R}^{\ell'_k \times C}$.


\subsubsection{Temporal Sampling}\label{sample}

Although it is possible to obtain $a'_k$ by just copying $\ell'_k$ times the probabilities $a_k$, this operation is not differentiable with respect to $\ell'_k$. However, we need a differentiable operation in order to update the parameters of $f_l$, which predicts $L$, during training. 

We first generate our target matrix $a'_k \in \mathbb{R}^{H \times C}$ where $H = \max_k \ell'_k$, \ie, the matrix is set such that the size is constant for all $k$. For a better temporal sampling, we also expand the source by copying $j$ times $a_k$, where $J$ is a canonical value equal to $100$. Although $a_k$ has been expanded to $\mathbb{R}^{J \times C}$, we still keep the notation $a_k$.    

The idea is to fill the matrix $a'_k$ by backward warping and a bilinear kernel. Similar to \cite{STN}, we use normalized element indices, such that $-1\leq i_a[j]\leq1$ when $j \in [1\dots J]$ and $-1\leq i_{a'}[h]\leq 1$ when $h \in [1\dots H]$. This means if we just interpolate the values for each column $c$, the operation is defined by  
\begin{equation}
\label{eq:inter}
    a'_k[h,c]= \sum_{j=1}^{J} a_k[j,c] \max\left(0,1-\left|i_{a'}[h]-i_a[j]\right|\right) 
\end{equation}
for $h \in [1\dots H]$. 

However, we do not want to fill the entire row but only until $\ell'_k$. We therefore apply a 1D affine transformation to the index function       
\begin{equation}
    \label{eq:affine_transformation}
    T_{\ell'_k}(i_{a'}[h]) = \frac{H}{\ell'_k} i_{a'}[h] + \frac{H}{\ell'_k} - 1 .
\end{equation}
This means that $T_{\ell'_k}(i_{a'}[1]) = -1$ and $T_{\ell'_k}(i_{a'}[\ell'_k]) = 1$.   
By integrating \eqref{eq:affine_transformation} into \eqref{eq:inter}, we obtain the upsample operation  
\begin{equation}
    a'_k[h,c]= \sum_{j=1}^{J} a_k[j,c] \max\left(0,1-\left|T_{\ell'_k}(i_{a'}[h])-i_a[j]\right|\right) 
\end{equation}
that is differentiable with respect to $\ell'_k$. 

Finally, the matrix is cropped to $a'_k \in \mathbb{R}^{\ell'_k \times C}$ and we obtain $Y \in \mathbb{R}^{T \times C}$ by concatenating the $a'_k$s for $k=1,\dots, K$.

%% file: Paper/5ProposedMethodTraining.tex
\section{Training}\label{sec:training}
In Section \ref{sec:propsedMethod} we proposed a model that is capable of dividing a temporal sequence into temporal regions and predicting corresponding action probabilities $A$ and lengths $L$. We now discuss the loss functions and regularizers for training the model. 

\subsection{Set Loss}\label{sec:setLoss}
In a set supervised problem we already have the set supervision. So we use a simple set prediction loss $\mathcal{L_{S}}$ to use the given set of actions $\hat{A}$. We apply a global max pooling over the temporal dimension of $A$ to output $a^{mc} \in \mathbb{R}^{C}$. Then we use the binary cross entropy loss for multiclass classification as

\begin{equation}
    \label{eq:setLoss}
    \mathcal{L_{S}}=  -\frac{1}{C}\left(\sum_{m \in \hat{A}} \log\left(a^{mc}[m]\right) + \sum_{m \notin \hat{A}} \log\left(1-a^{mc}[m]\right)\right).
\end{equation}
This loss encourages the model to assign at least one region to one of the classes in $\hat{A}$ and none to the other classes.    

\subsection{Region Loss}
The set loss only enforces that there is one region with a high probability for each class. It can, however, happen that the other regions have a uniform distribution for the classes in $\hat{A}$.    
Since it is unlikely that all actions occur at the same time, we introduce the region loss, which encourages the model to predict only one action from $\hat{A}$ per region. Since we know that only actions from $\hat{A}$ can occur, we first discard the unrelated actions from $A \in \mathbb{R}^{K \times C}$ and denote it by $A^{\mathcal{S}} \in \mathbb{R}^{K \times M}$, where each column belongs to one of the given action set members $\hat{a}_m$. We now prefer a prediction where for each $k$ the probability is close to 1 for one action $\hat{a}_m$. Due to the softmax, this means that the probability is close to zero for the other actions. 

This is achieved by applying a global max pooling over the $m$ dimension of $A^{\mathcal{S}} \in \mathbb{R}^{K \times M}$ to obtain $a^{mk} \in \mathbb{R}^{K}$ and using the cross entropy loss: 
\begin{equation}
    \label{eq:setLoss}
    \mathcal{L_{R}}=  - \frac{1}{K}\sum_{k=1}^{K} \log\left(a^{mk}[k]\right).
\end{equation}



\subsection{Inverse Sparsity Regularization}\label{sec:invSparsityReg}

The set loss and the region loss ensure that (i) all actions that are not in the set $\hat{A}$ have a low probability, (ii) for each temporal region there is exactly one action $\hat{a}_m \in \hat{A}$ with high probability, and (iii) for each action $\hat{a}_m$ there is at least one region $k$ where $a[k,m]$ is high. This, however, can result in unlikely solutions where for $M-1$ actions there is only one region with high probability whereas the other regions are assigned to a single action class. To prevent such a sparse distribution of regions for some action classes, we introduce an inverse sparsity regularization term $\mathcal{R_{I}}$, which prefers a more balanced class distribution averaged over all regions:  
\begin{equation}
    \label{eq:invSparsityReg}
    \mathcal{R_{I}}= \frac{1}{M}\sum_{m\in\hat{A}}\left(1 - \frac{1}{K}\sum_{k=1}^{K}a[k,m]\right).
\end{equation}
This regularizer encourages that the action classes compete for maximizing the number of temporal regions they are being predicted for.

\subsection{Temporal Consistency Loss}\label{sec:temporalConsistencyLoss}
As illustrated in Figure~\ref{fig:temporalRegionsLengths}, the temporal regions are usually smaller than the action segments in the video and a single action often spans several regions. The likelihood of observing the same action in the neighboring temporal regions is therefore usually higher than observing a different action. We therefore introduce the temporal consistency loss $\mathcal{L_{C}}$, that encourages the model to predict similar action labels for neighboring temporal regions:  
\begin{equation}
    \label{eq:tmpConsistencyLoss}
    \mathcal{L_{C}}= \frac{1}{M}\sum_{m\in\hat{A}}\frac{1}{K}\sum_{k=2}^{K}|a[k,m]-a[k-1,m]|.
\end{equation}
More precisely, $\mathcal{L_{C}}$ encourages the model to have less prediction changes over the temporal dimension of $A^{S}$.

\subsection{Self Supervision Loss}\label{sec:selfSupLoss}

The aforementioned losses and regularizers only affect the class probabilities $A$ of the temporal regions, but do not backpropagate gradients through the subnetwork $f_{l}$. This means that the network does not learn the corresponding lengths of the regions during training. In order to provide an auxiliary supervision signal to train $L$, we employ a self supervision technique which relies on using two different representations. The first representation $Y$ is obtained by estimating the actions probabilities $A$ and lengths $L$ for $K$ temporal regions as described in Section~\ref{sec:tempSegment}. Due to the temporal sampling, the representation $Y$ is differentiable with respect to $L$. 

To have another representation, we use a second branch $f_s(Z)$. This branch consists of a subnetwork that has a single 1-d convolution with kernel size $1$ and output channel size $C$. It predicts frame-wise class probabilities for the temporal size $T'$. We linearly interpolate it to $S \in \mathbb{R}^{T \times C}$ along the temporal dimension, which corresponds to a setting where $K=T'$ and $\ell_k = \frac{T}{T'}$, \ie, all regions have a constant length. 

Since we do not know the ground-truth lengths $L$ but only the set of present actions $\hat{A}$, we combine $Y$ and $S$ to compute class probabilities $V_{1:M}=(v_1,...v_M),v_m \in \mathbb{R}^{C}$, for each element in the set $\hat{A}$. This is done by the Set Constrained Temporal Transformer module (SCT). 



\subsection{Set Constrained Temporal Transformer}

As it is illustrated in Figure~\ref{fig:overview}, we produce for each action class $\hat{a}_m \in \hat{A}$ masks $w_m$ from $Y$. The masks indicate the temporal locations where the action $\hat{a}_m$ occurs in the video. We use these masks to sample from $S$: 
\begin{equation}
    \label{eq:samplingOverS}
    v_m = \frac{1}{T}\sum_{t=1}^{T}w_m[t]S[t].
\end{equation}
If $S$ and $Y$ are consistent, $v_m[\hat{a}_m]$ should be high and $v_m[\hat{a}_n]$ should be close to zero for $n \neq m$. 

To exploit this, we apply a softmax on $v_m$ to get the predicted probabilities for the given action and use the cross entropy loss: 
\begin{equation}
    \label{eq:actionMaskLoss}
    \mathcal{L}_{\mathcal{T}_m}(v_m, \hat{a}_m) = -\log\left(\frac{e^{v_{m}[\hat{a}_m]}}{\sum_{c=1}^{C}e^{v_{m}[c]}}\right).
\end{equation}  
Since $w_m$ is differentiable with respect to $a_m$ and $l_m$, the loss affects both.  

As a more efficient way, we can apply all of the masks $W$ on $S$ using:
\begin{equation}
    \label{eq:transposedMatMul}
    V = \frac{W^TS}{T}
\end{equation}
where $V \in \mathbb{R}^{M \times C}$ and $W^T\in \mathbb{R}^{M \times T}$ denotes the transposed version of $W$. Therefore, we can define the loss for $V$ and the given actions set $\hat{A}$ as
\begin{equation}
    \label{eq:invTempTrnLoss}
    \mathcal{L}_{\mathcal{T}}(V, \hat{A}) = -\frac{1}{M}\sum_{m=1}^{M} \log\left(\frac{e^{V[m,\hat{a}_m]}}{\sum_{c=1}^{C} e^{V[m,c]}}\right).
\end{equation}

\subsubsection{Backpropagation}

Using the $\mathcal{L}_{\mathcal{T}}$ loss, the gradient can backpropagate through both $S$ and $Y$. $Y$ is the output of $f_u(A,L)$ which is a differential function over $L$ and $A$. Therefore, we can update the $f_{l}$ weights using the backpropagated gradients. To be able to backpropagate through the $a'_k$s we define the gradients with respect to the sampling indices $T_{\ell'_k}(i_{a'}[h])$ as
\begin{align}
    \label{eq:differentiable_sampling_bilinear}
    &\frac{\partial a'_k[h,c]}{\partial T_{\ell'_k}(i_{a'}[h])}=\nonumber\\
    &\quad\sum_{j=1}^{J}a_k[j,c] \: 
    \begin{cases}
    0 & |i_a[j]-T_{\ell'_k}(i_{a'}[h])|\geq1 \\ 
    1 & i_a[j]-1< T_{\ell'_k}(i_{a'}[h]) \leq i_a[j] \\ 
    -1 & i_a[j]<T_{\ell'_k}(i_{a'}[h])<i_a[j]+1
    \end{cases}.
\end{align}
Since the sampling indices $T_{\ell'_k}(i_{a'}[h])$ are a function of the predicted lengths $L_{1:M}$, the loss gradients are backpropagated to the predicted lengths

\subsubsection{Region Length Regularization}\label{sec:regionLengthRegularization}
Learning the lengths based on $\mathcal{L}_{\mathcal{T}}$ may result in degenerated lengths which are close to zero. Therefore, we use a length regularizer $\mathcal{R_{L}}$ to prevent such circumstances. We define $\mathcal{R_{L}}$ as
\begin{equation}
    \label{eq:lengthRegularizer}
    \mathcal{R_{L}} = \frac{1}{K}\sum_{t=1}^{K}(ReLU(\ell_{t} - \delta) + ReLU(-\ell_{t} - \delta))
\end{equation}
where $\delta$ is a canonical value equal to $1$. This regularization term penalizes the lengths which are bigger or smaller than the length width of $\delta$.

\subsection{Overall Loss}
All of the loss functions and regularizers that we mentioned in this section encourage the model to exploit the given weak supervision and also characteristics of actions to train the model for action segmentation. Therefore, the final loss function for the model is the weighted sum of the above mentioned losses and regularizers. In Section \ref{sec:experiments} we study the impact of all loss functions and regularizers.

%% file: Paper/6Experiments.tex
\section{Experiments} \label{sec:experiments}
\begin{figure*}[t]
    \centering
    \includegraphics[scale=0.25,bb=0 0 2000 700]{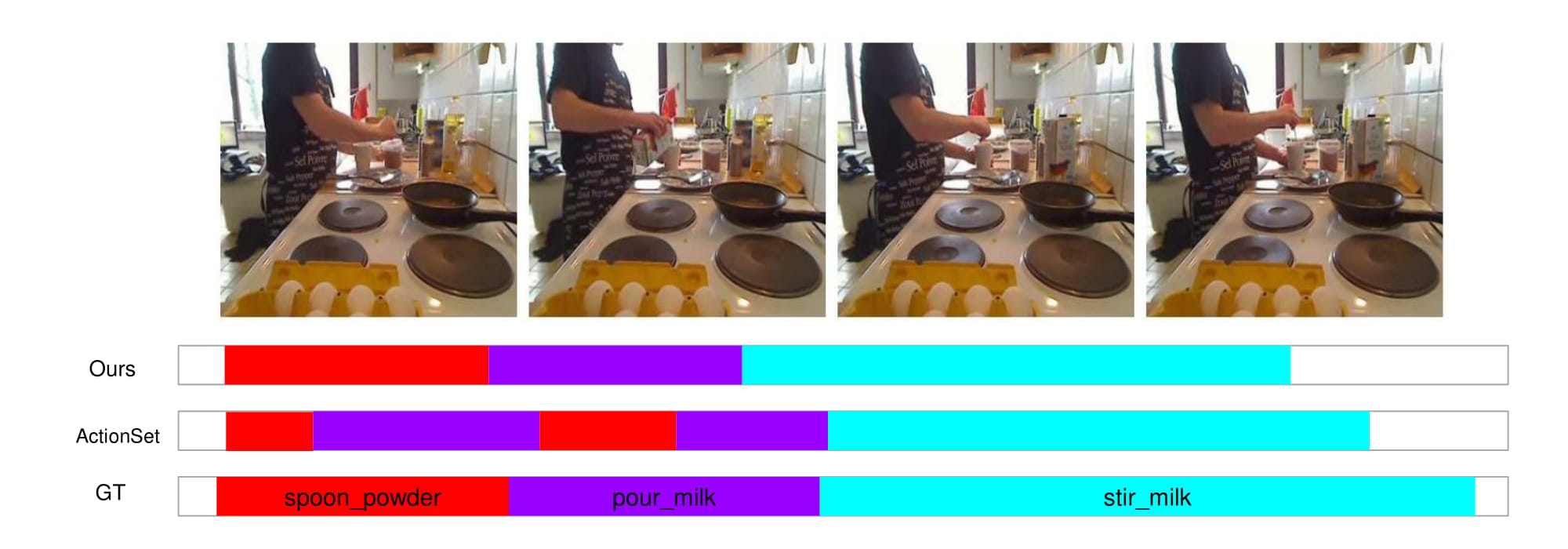}
    \caption{Comparing the segmentation quality of our method to the ActionSet method. Our method has predicted the right order of actions occurring in this video. Our approach also estimates the actions lengths better. }
    \label{fig:qualitativeResults}
\end{figure*}
In this section, we analyze the components of our approach. We first analyze the model design. Then we evaluate the effect of using different loss functions and regularizers. Finally, we compare our method with the state-of-the-art.
\subsection{Setup}
\textbf{Datasets.}
We evaluate our method on three popular datasets, namely the Breakfast dataset~\cite{BreakFastDataset}, Hollywood Extended~\cite{hollywoodextended}, and MPII Cooking 2~\cite{MPII2}.

The \textbf{Breakfast} dataset contains $1,712$ videos of different cooking activities, corresponding to about $67$ hours of
videos and $3.6$ million frames. The videos belong to $10$ different types of breakfast activities like \textit{fried egg} or \textit{coffee} which consist of $48$ different fine-grained actions. The actions are densely annotated
and only $7\%$ of the frames are background. We report the average frame accuracy (MoF) metric over the predefined train/test splits following \cite{ActionSet}.

\textbf{Hollywood Extended} contains $937$ video sequences with roughly $800,000$ frames. About $61\%$ of the frames are background, which is comparably large compared to other datasets. The videos contain $16$ different action classes. We report the Jaccard index (intersection over union) metric over the predefined train/test splits following \cite{ActionSet}.

\textbf{MPII 2 Cooking} consists of $273$ videos with about $2.8$ million frames. We use the $67$ action classes without object annotations. About $29\%$ of the frames of this dataset are background. The dataset provides a fixed split into a train and test set, separating 220 videos for training.  For evaluation, we use the midpoint hit criterion following~\cite{cookingActivitiesDataset}.

\textbf{Feature Extraction.}
We use RGB+flow I3D \cite{i3d} features extracted from the I3D network pretrained on the Kinetics400 dataset \cite{kinetics}. The I3D features are extracted for each frame. Moreover, for a fair comparison we also use the same IDT features as \cite{ActionSet} and evaluate the effect of using different features.

\textbf{Implementation Details.}
We train all modules of our network together. The hidden size of the temporal embedding module is $128$.
We use SGD optimizer with weight decay $0.005$. The initial learning rate is set to $0.01$. Additional details and code are available online.\footnote{http://mohsenfayyaz89.github.io}

\subsection{Ablation Experiments}
In this section we first analyze the model design. Then we analyze the effect of our loss functions and regularizers on training the model.
\subsubsection{Effect of different downsampling levels}
As mentioned in Section \ref{sec:temporalmodule}, we downsample the input by applying temporal max pooling with kernel size $2$. We evaluate the effect of downsampling by changing the numbers of temporal max pooling operations in the temporal embedding module. It should be mentioned that we apply each max pooling on top of each temporal convolution block (TCB). As can be seen in Table~\ref{table:maxpoolings}, a small number of max pooling operations results in a relatively low frame-wise accuracy. This is due to high number of temporal regions, which may result in an over-segmentation problem. Furthermore, a drop in performance can be observed when the number of max pooling operations is too large. In this case, there are not enough temporal regions and a temporal region covers multiple actions. For the rest of the experiments, we use $3$ max pooling operations in our temporal modeling module. It should be noted that we use $3$ max pooling operations after the TCBs with indices $\{1,2,4\}$, while in this experiment $3$ max pooling operations are applied after the TCBs with indices $\{1,2,3\}$.
\begin{table}[t]
\centering
\small
\tabcolsep=0.15cm
\begin{tabular}{ c| c c c c c c c}
  \#max poolings  & $0$ & $1$ & $2$ & $3$ & $4$ & $5$ & $6$ \\
  \hline 
  MoF & $12.3$ & $15.4$ & $20.8$ & $28.1$ & $27.2$ & $21.3$ & $18.2$ \\

\end{tabular}
\caption{Evaluating the effect of changing the numbers of max pooling operations in the temporal embedding module. Experiments are run on Breakfast split~$1$.}
\label{table:maxpoolings}
\end{table}

\subsubsection{Effect of using different loss functions and regularizers}
As mentioned in Section~\ref{sec:training}, we use different loss functions and regularizers to train our model. To quantify the effect of using these loss functions and regularizers, we train our model with different settings in which we can evaluate the effect of them for training. We train our model on split~$1$ of the Breakfast dataset and report the results in Table~\ref{table:LossRegularizersComparison}.

As mentioned in Section~\ref{sec:setLoss}, the \textbf{Set Loss} $\mathcal{L_{S}}$ encourages the model to have at least one temporal region with a high class probability for each action in the target action set. Therefore, this loss does not affect the frame-wise prediction performance of the model. As it can be seen in Table~\ref{table:LossRegularizersComparison}, using only $\mathcal{L_{S}}$ the model achieves MoF of $8.1\%$.

By adding the \textbf{Region Loss} $\mathcal{L_{R}}$, we encourage the model to only predict one action per temporal region. Using this auxiliary supervision, the MoF slightly improves to $9.9\%$ which is still relatively low.

As mentioned in Section~\ref{sec:invSparsityReg}, adding the \textbf{Inverse Sparsity Regularizer} $\mathcal{R_{I}}$ helps the method to prevent a sparse distribution of regions for some action classes. Therefore, by adding $\mathcal{R_{I}}$ to the overall loss, the MoF improves to $19.2\%$, which is significantly better than predicting every frame as background which covers about $7\%$ of the frames.

We further add the \textbf{Temporal Consistency Loss} $\mathcal{L_{C}}$ to encourage the model to predict similar actions for neighboring temporal regions. $\mathcal{L_{C}}$ improves the result to $21.9\%$.

As mentioned in Section~\ref{sec:selfSupLoss}, all of the aforementioned losses and regularizers only affect the class probabilities $A$ of the temporal regions and do not backpropagate gradients through the length estimator head $f_{l}$. We therefore add the \textbf{Self Supervision Loss} $\mathcal{L_{T}}$ to evaluate the effect of learning lengths during training. Adding $\mathcal{L_{T}}$ significantly improves the accuracy to $29.9\%$. This improvement shows the effect of refining the temporal regions using the predicted lengths.

We also evaluate the effect of using the \textbf{Region Length Regularization} $\mathcal{R}_{\mathcal{L}}$. As mentioned in Section~\ref{sec:regionLengthRegularization}, learning the lengths only based on $\mathcal{L}_{\mathcal{T}}$ may result in too diverse estimated lengths for temporal regions. Therefore, we evaluate the effect of $\mathcal{R}_{\mathcal{L}}$ by adding it to the overall loss. By adding this regularizer the accuracy improves to $30.8\%$. Since $\mathcal{L}_{\mathcal{T}}$ and $\mathcal{R}_{\mathcal{L}}$ are the only loss function and regularizer which affect the lengths $L$, we also evaluate the effect of only using $\mathcal{R}_{\mathcal{L}}$ as an effective regularizer on the lengths without $\mathcal{L}_{\mathcal{T}}$. This setting results in an MoF of $22.2\%$. The reason for such a significant drop in performance is that $\mathcal{R}_{\mathcal{L}}$ only encourages the model to not estimate too diverse lengths. This shows that the proposed self supervision loss based on the set constrained temporal transformer is important to learn proper lengths for the temporal regions.

To have a better understanding of the \textbf{Self Supervision Loss}, we also try to train the temporal regions' length estimator head $f_l$ in a different way. Instead of using $\mathcal{L_T}$, we use the Jensen Shannon Divergence loss which is a symmetric and smoothed version of the Kullback–Leibler divergence to match both representations $Y$ and $S$ as follows:
\begin{equation}
    \label{eq:JSDLoss}
    \mathcal{L_{J}} = \frac{1}{2}D(Y\parallel M)+D(S\parallel M),
\end{equation}
\begin{equation}
    \label{eq:JSDLoss}
    \mathcal{D(P \parallel Q)} = \sum_{x\in X}P(x)\log(\frac{P(x)}{Q(x)})
\end{equation}
where $M = \frac{1}{2}(Y + S)$. Using $\mathcal{L_{J}}$ instead of $\mathcal{L_{T}}$ results in an MoF of $25.3\%$, which shows the superiority of our self supervision loss.
\
\begin{table}[t]
\centering
\small
\tabcolsep=0.15cm
\begin{tabular}{ c c c c c c c | l}
  $\mathcal{L_{S}}$ & $\mathcal{L_{R}}$ & $\mathcal{R_{I}}$ & $\mathcal{L_{C}}$ & $\mathcal{L_{T}}$ & $\mathcal{R_{L}}$ & $\mathcal{L_{J}}$ & MoF \\
\hline 
\hline 
 \checkmark & - & - & - & - & - & - & 8.1\\
 \hline
 \checkmark & \checkmark & - & - & - & - & - & 9.9\\
 \hline
 \checkmark & \checkmark & \checkmark & - & - & - & - & 19.2\\
 \hline
 \checkmark & \checkmark & \checkmark & \checkmark & - & - & - & 21.9\\
 \hline
 \checkmark & \checkmark & \checkmark & \checkmark & \checkmark & - & - & 29.9\\
 \hline
 \checkmark & \checkmark & \checkmark & \checkmark & \checkmark & \checkmark & - & \textbf{30.8}\\
 \hline
 \checkmark & \checkmark & \checkmark & \checkmark & - & \checkmark & - & 22.2\\
 \hline
 \checkmark & \checkmark & \checkmark & \checkmark & - & \checkmark & \checkmark & 25.3\\

\end{tabular}
\caption{Evaluating the effect of using different losses and regularizers. Experiments are run on Breakfast split~$1$.}
\label{table:LossRegularizersComparison}
\end{table}

\subsubsection{Effect of using different features}
As mentioned before, we use I3D features \cite{i3d} as input to our model. To evaluate the effect of the input video features, we train our model using IDT features as well and report the results in Table~\ref{table:FeatursComparison}. To have a better comparison with the previous state-of-the-art method~\cite{ActionSet}, we also train this method with I3D features using the publicly available code. We observe that our method achieves state-of-the-art results using both types of features. The ActionSet~\cite{ActionSet} method does not perform well on I3D features which may be due to the limitations in its temporal architecture design that is not capable of learning a proper temporal embedding over the I3D features.
\begin{table}[t]
\centering
\small
\tabcolsep=0.15cm
\begin{tabular}{ l|l c }
   & I3D & IDT \\
\hline 
\hline 
ActionSet \cite{ActionSet}  & 20.1* &  23.3\\
\textbf{Ours} & \textbf{30.4} & \textbf{26.6} \\
\end{tabular}
\caption{Comparison of our method to \cite{ActionSet} for different features. Experiments are run on the Breakfast dataset and MoF is reported. Our method achieves state-of-the-art results using both types of features. *The source code of the paper has been used for this experiment.}
\label{table:FeatursComparison}
\end{table}

\subsection{Comparison to State-of-the-Art}
The task of learning temporal action segmentation using action sets as weak supervision has been so far addressed only by~\cite{ActionSet}.
We compare our approach to this method on three datasets. As it can be seen in Table~\ref{table:SOAComparison}, our method achieves state-of-the-art results on all three datasets. While the methods with only set supervision work well on Breakfast and Hollywood Extended, the performance on the Cooking 2 dataset is lower. The Cooking 2 dataset has a high number of classes $(67)$ while having a low number of training samples $(220)$. The other problem is that this dataset contains very long videos which on average contain $50$ different actions. Therefore, learning a temporal action segmentation model on this dataset using only weak set supervision is very difficult. In order to get a better understanding of the effect of such characteristics of this dataset, we have evaluated the effect of cutting this dataset into different parts followed by~\cite{ActionSet}. As it can be seen in Table~\ref{table:cookingCuts}, having more videos for training and fewer actions per video on average improves the results. Figure~\ref{fig:qualitativeResults} shows a qualitative result of our method for a video from the Breakfast dataset.

\begin{table}[t]
\centering
\small
\tabcolsep=0.15cm
\begin{tabular}{ l|c c c }
 Dataset  & Break Fast & Holl. Ext. & Cooking 2 \\
  & \textit{MoF} & \textit{jacc. idx} & \textit{midpoint hit}\\
\hline 
\hline 
ActionSet-monte-carlo \cite{ActionSet}  & 23.3 &  9.3& 9.8\\
ActionSet-text-based \cite{ActionSet}  &  23.2& 9.2& 10.6\\
Ours & \textbf{30.4} & \textbf{17.7} & \textbf{14.3}\\
\end{tabular}
\caption{Comparison of our method to \cite{ActionSet} for weakly supervised temporal segmentation.}
\label{table:SOAComparison}
\end{table}

\begin{table}[t]
\centering
\small
\tabcolsep=0.15cm
\begin{tabular}{ l | c c c }
 cuts per video  & 4 & 2 & - \\
\hline 
\hline 
avg. \# actions per video & 12.5 & 25 & 50\\
\hline 
ActionSet \cite{ActionSet}  & 17.4 &  12.1& 9.8\\
Ours & \textbf{19.8} & \textbf{16.1} & \textbf{14.3}\\
\end{tabular}
\caption{Different levels of video trimming for Cooking 2. More
videos and fewer actions per video result in better performance.}
\label{table:cookingCuts}
\end{table}

%% file: Paper/7Concolusion.tex
\section{Conclusion}
In this work we presented a network for temporal action segmentation. The network is trained on long videos which are only annotated by the set of present actions. The network is trained by dividing the videos into temporal regions that contain only one action class and are consistent with the set of annotated actions. We thoroughly evaluated the approach on three datasets. For all three datasets, the proposed network outperforms previous work.            

%% file: Paper/8Acknowledgment.tex
\paragraph{Acknowledgment}
The work has been funded by the Deutsche Forschungsgemeinschaft (DFG, German Research Foundation) GA 1927/4-1 (FOR 2535 Anticipating Human Behavior) and the ERC Starting Grant ARCA (677650).